\documentclass[10pt,journal,compsoc]{IEEEtran}
\ifCLASSOPTIONcompsoc
  \usepackage[nocompress]{cite}
 \usepackage{amsmath,amssymb,amsfonts}
\usepackage{bbm}
\usepackage{graphicx}
\usepackage{textcomp}
\usepackage{xcolor}
\usepackage{amsmath}
\usepackage{multirow}
\usepackage{ulem}
\usepackage{url}
\usepackage{hyperref}
\usepackage{multirow}
\usepackage{ulem}
\usepackage{url}
\usepackage{booktabs}
\usepackage{subfigure}
\usepackage{amssymb}
\usepackage{graphicx}
\usepackage{array}
\usepackage{amssymb}
\usepackage{amsmath}
\usepackage{threeparttable}
\usepackage{color,xcolor}
\usepackage{ulem}
\usepackage{algorithm}
\usepackage{algorithmicx}
\usepackage{algpseudocode}
\usepackage{ulem}

\else
  \usepackage{cite}
\fi

\ifCLASSINFOpdf
\else
\fi
\begin{document}
\title{An Efficient Federated Distillation Learning System for Multi-task Time Series Classification}

\author{
        Huanlai~Xing, ~\IEEEmembership{Member,~IEEE}, ~Zhiwen~Xiao,  ~Rong~Qu,~\IEEEmembership{Senior~Member,~IEEE}, ~Zonghai~Zhu, and~Bowen~Zhao
        % <-this % stops a space
\IEEEcompsocitemizethanks{\IEEEcompsocthanksitem  H. Xing, Z. Zhu, and B. Zhao are with the School of Computing and Artificial Intelligence, Southwest Jiaotong University, Chengdu 611756, China (Emails: hxx@home.swjtu.edu.cn; zzhu@swjtu.edu.cn; cn16bz@icloud.com).

\IEEEcompsocthanksitem Z. Xiao is with Southwest Jiaotong University, Chengdu 611756, China, and Chengdu University of Information Technology, Chengdu 610103, China  (Email: xiao1994zw@163.com).

\IEEEcompsocthanksitem R. Qu is with the the School of Computer Science, University of Nottingham, Nottingham NG7 2RD 455356, UK (Email: rong.qu@nottingham.ac.uk)}% <-this % stops an unwanted space
\thanks{Manuscript received XX, XX; revised  XX, XX (Corresponding author: Zhiwen Xiao).}}


\markboth{Journal of \LaTeX\ Class Files,~Vol.~14, No.~8, August~2015}%
{Shell \MakeLowercase{\textit{et al.}}: Bare Demo of IEEEtran.cls for Computer Society Journals}

\IEEEtitleabstractindextext{%
\begin{abstract}
This paper proposes an efficient federated distillation learning system (EFDLS) for multi-task time series classification (TSC). EFDLS consists of a central server and multiple mobile users, where different users may run different TSC tasks. EFDLS has two novel components, namely a feature-based student-teacher (FBST) framework and a distance-based weights matching (DBWM) scheme. Within each user, the FBST framework transfers knowledge from its teacher's hidden layers to its student's hidden layers via knowledge distillation, with the teacher and student having identical network structure. For each connected user, its student model's hidden layers' weights are uploaded to the EFDLS server periodically. The DBWM scheme is deployed on the server, with the least square distance used to measure the similarity between the weights of two given models. This scheme finds a partner for each connected user such that the user's and its partner's weights are the closest among all the weights uploaded. The server exchanges and sends back the user's and its partner's weights to these two users which then load the received weights to their teachers' hidden layers. Experimental results show that the proposed EFDLS achieves excellent performance on a set of selected UCR2018 datasets regarding top-1 accuracy.
\end{abstract}

}

%Each user runs one TSC task at a time and different users may run different TSC tasks (e.g., gesture and emotion classifications). 
\maketitle

\IEEEdisplaynontitleabstractindextext

% \fi
\IEEEpeerreviewmaketitle

\IEEEraisesectionheading{\section{Introduction}\label{sec:introduction}}
\IEEEPARstart{T}{ime} series data is a series of time-ordered data points associated with one or more time-dependent variables and has been successfully applied to areas such as anomaly detection \cite{F1,F5}, traffic flow forecasting \cite{F2}, service matching \cite{F4}, stock prediction \cite{F6}, electroencephalogram (ECG) detection \cite{F7} and parking behavior prediction \cite{d1}. A significant amount of research attention has been dedicated to time series classification (TSC) \cite{F8}. For example, Wang \textit{et al.} \cite{F9} introduced a fully convolutional network (FCN) for local feature extraction. Zhang \textit{et al.} \cite{F10} devised an attentional prototype network (TapNet) to capture rich representations from the input. Karim \textit{et al.} \cite{F11} proposed a long short-term memory (LSTM) fully convolutional network (FCN-LSTM) for multivariate TSC. A robust temporal feature network (RTFN) hybridizing temporal feature network and LSTM-based attention network was applied to extracting both the local and global patterns of data \cite{F12}. Li \textit{et al.} \cite{F13} put forward a shapelet-neural network approach to mine highly-diversified representative shapelets from the input. Lee \textit{et al.} \cite{F14} designed a learnable dynamic temporal pooling method to reduce the temporal pooling size of the hidden representations obtained.

TSC algorithms are usually data-driven, where data comes from various application domains. Some data may contain private and sensitive information, such as bank account and ECG. However, traditional data collection operations could not well protect such information, easily resulting in users' privacy leakage during the data collection and distribution processes involved in model training. To overcome the problem above, Google \cite{F15,F16,F41} invented federated learning (FL). FL allows users to collectively harvest the advantages of shared models trained from their local data without sending original data to others. FederatedAveraging (FedAvg), federated transfer learning (FTL) and federated knowledge distillation (FKD) are the three mainstream research directions.

FedAvg calculates the average weights of the models of all users and shares the weights with each user in the FL system \cite{F17}. For instance, Ma \textit{et al.} \cite{F19} devised a communication-efficient federated generalized tensor factorization for electronic health records. Liu \textit{et al.} \cite{F20} used a federated adaptation framework to leverage the sparsity property of neural networks for generating privacy-preserving representations. A hierarchical personalized FL method aggregated heterogeneous user models, with privacy heterogeneity and model heterogeneity considered \cite{F21}. Yang \textit{et al.} \cite{b1} modified the FedAvg method using partial networks for COVID-19 detection. 

FTL introduces transfer learning techniques to promote knowledge transfer between different users, increasing system accuracy \cite{F22}. For example, Yang \textit{et al.} \cite{F32} developed an FTL framework, FedSteg, for secure image steganalysis. An FTL method with dynamic gradient aggregation was proposed to weight the local gradients during the aggregation step when handling speech recognition tasks \cite{F33}. Majeed \textit{et al.} \cite{b20} proposed an FTL-based structure to address traffic classification problems.

Unlike FedAvg and FTL, FKD takes the average of all users' weights as the weights for all teachers and transfers each teacher's knowledge to its corresponding student via knowledge distillation (KD) \cite{F25}. A group knowledge transfer training algorithm was adopted to train small convolutional neural networks (CNNs) and transfer their knowledge to a prominent server-side CNN \cite{F26}. Mishra \textit{et al.} \cite{b21} proposed a resource-aware FKD approach for network resource allocation. Sohei \textit{et al.} \cite{F43} devised a distillation-based semi-supervised FL framework for communication-efficient collaborative training with private data. Nowadays, FKD is attracting increasingly more research attention.

In addition, there is a variety of FL-based algorithms in the literature. For instance, Chen \textit{et al.} \cite{F23} applied asynchronous learning and temporally weighted aggregation to enhancing system's performance. Sattler \textit{et al.} \cite{F24} presented a sparse ternary compression method to meet various requirements of FL environment. A cooperative game involving a gradient algorithm was designed to tackle image classification and speech recognition tasks \cite{F27}. An ensemble FL system used a randomly selected subset of clients to learn multiple global models against malicious clients \cite{F28}. Hong \textit{et al.} \cite{b2} combined adversarial learning and FL to produce federated adversarial debiasing for fair and transferable representations. Zhou \textit{et al.} \cite{F40} proposed a privacy-preserving distributed contextual federated online learning framework with big data support for social recommender systems. Pan \textit{et al.} \cite{F42} put forward a multi-granular federated neural architecture search framework to enable the automation of model architecture search in a federated and privacy-preserved setting.

Most FL algorithms are developed around single-task problems, where multiple users work together to complete a task, e.g., COVID-19 detection \cite{b1}, traffic classification \cite{b20} or speech recognition \cite{F33}. It is quite challenging to directly apply these algorithms to multi-task problems unless efficient knowledge sharing among different tasks is enabled. Unfortunately, TSC is usually multi-task-oriented. Time series data is collected from various application domains, such as ECG, traffic flow, human activity recognition. Each time series dataset has specific characteristics, e.g., length and variance, which may differ significantly from others. Thus, time series data is highly imbalanced and strongly non-independent, and identically distributed (Non-I.I.D.). In multi-task learning, it is commonly recognized that knowledge sharing among different tasks helps increase the efficiency and accuracy of each task \cite{F3}. \textit{For most TSC algorithms, how to securely share knowledge of similar expertise among different tasks is still challenging. In other words, user privacy and knowledge sharing are two critical issues that need to be carefully addressed when devising practical multi-task TSC algorithms}. To the best of our knowledge, FL for multi-task TSC has not received sufficient research attention.

We present an efficient federated distillation learning system (EFDLS) for multi-task TSC. This system consists of a central server and a number of mobile users running various TSC tasks simultaneously. Given two arbitrary users, they run either different tasks (e.g., ECG and motion) or the same task with different data sources to mimic real-world applications. EFDLS is characterized by a feature-based student-teacher (FBST) framework and a distance-based weights matching (DBWM) scheme. The FBST framework is deployed on each user, where the student and teacher models have identical network structure. Within each user, its teacher's hidden layers' knowledge is transferred to its student's hidden layers, helping the student mine high-quality features from the data. The DBWM scheme is deployed on the EFDLS server, where the least square distance (LSD) is used to measure the similarity between the weights of two models. When all connected users' weights are uploaded completely, for an arbitrary connected user, the DBWM scheme finds the one with the most similar weights among all connected users. After that, the server sends the connected user's weights to the found one that then loads the weights to its teacher model's hidden layers.

Our main contributions are summarized below.

\begin{itemize}

\item We propose EFDLS for multi-task TSC, where each user runs one TSC task at a time and different users may run different TSC tasks. The data generated on different users is different. EFDLS aims at providing secure knowledge sharing of similar expertise among different tasks. This problem has not attracted enough research attention.

\item In EFDLS, feature-based knowledge distillation is used for knowledge transfer within each user. Unlike the traditional FKD that adopts the average weights of all users to supervise the feature extraction process in each user, EFDLS finds the one with the most similar expertise (i.e., a partner) for each user according to LSD and offers knowledge sharing between the user and its partner.

\item Experimental results demonstrate that EFDLS outperforms six state-of-the-art FL algorithms considering 44 well-known datasets selected in the UCR 2018 archive regarding the mean accuracy, `win’/`tie’/`lose’ measure, and AVG\_rank, which are all based on the top-1 accuracy. That shows the effectiveness of EFDLS in addressing TSC problems.
\end{itemize}

The rest of the paper is organized below. Section 2 reviews the existing TSC algorithms. Section 3 overviews the architecture of EFDLS and describes its key components. Section 4 provides and analyzes the experimental results, and conclusion is drawn in Section 5.

\begin{figure*}[t]
\label{figg1}
  \centering
\includegraphics[width=18.2cm]{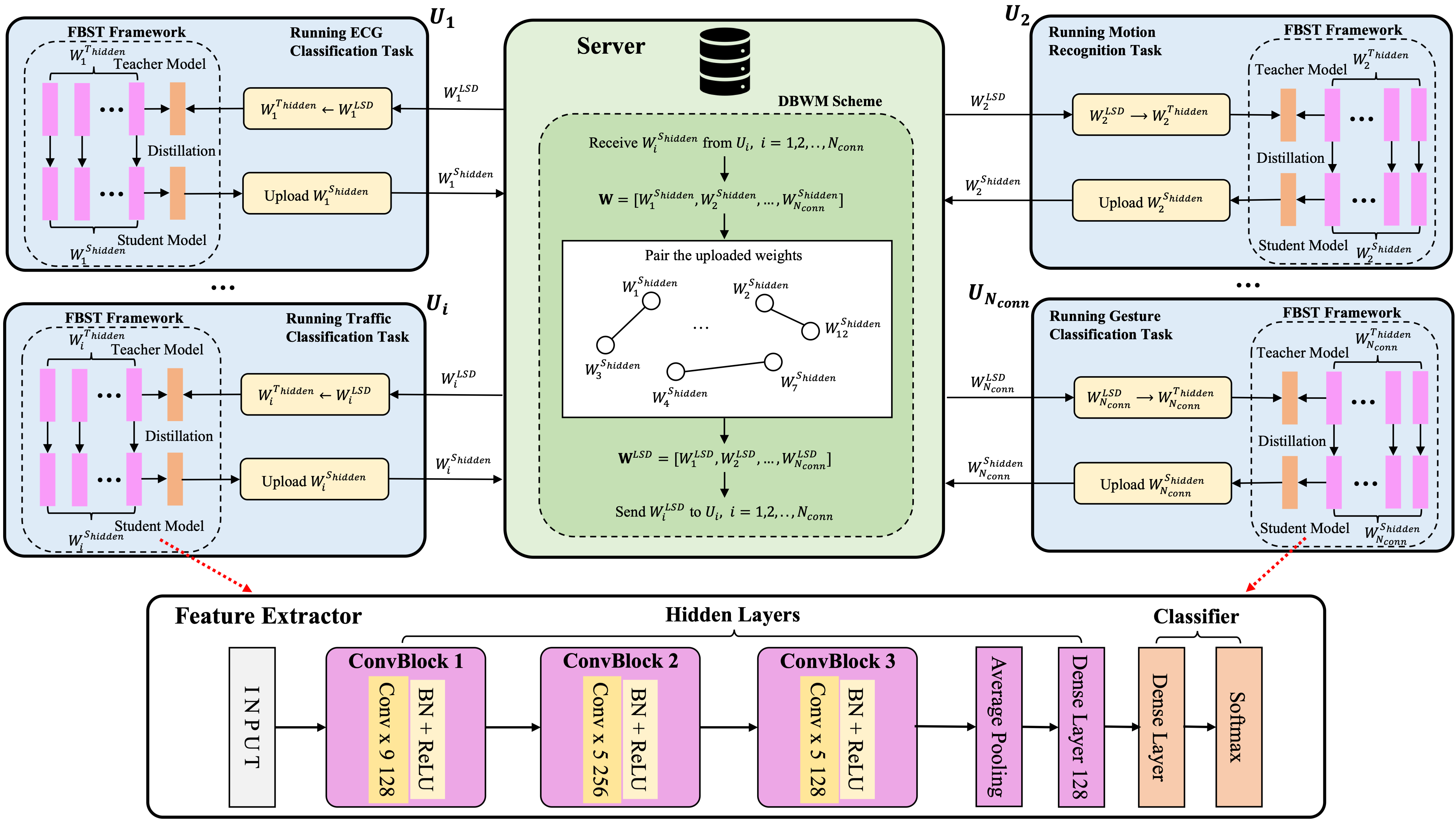}
\caption{The schematic diagram of EFDLS. Note that `FBST Framework' and `DBWM Scheme' denote the feature-based student-teacher framework deployed on each user and the distance-based weights matching scheme run on the server. `Conv x 9 128’ represents a 1-dimensional convolutional neural network, where its filter size and channel sizes are set to 9 and 128. `BN’ is a batch normalization module, and `ReLU’ is the rectified linear unit activation function.}
\end{figure*}

\section{Related Work}
A large number of traditional and deep learning algorithms have been developed for TSC. 
\subsection{Traditional Algorithms}
Two representative streams of algorithms are distance- and feature-based. For distance-based algorithms, it is quite common to combine the dynamic time warping (DTW) and nearest neighbor (NN), e.g., $DTW_{A}$, $DTW_{I}$ and $DTW_{D}$ \cite{b3}. Besides, a significant number of DTW-NN-based ensemble algorithms taking advantage of DTW and NN have been proposed in the community. For example, Line \textit{et al.} \cite{b4} presented an elastic ensemble (EE) algorithm for feature extraction, with 11 types of 1-NN-based elastic distance considered. A collective of the transformation-based ensemble (COTE) with 37 NN-based classifiers was adopted to address various TSC problems \cite{b5}. The hierarchical vote collective of transformation-based ensembles (HIVE-COTE) \cite{b6} and local cascade ensemble \cite{b7} are two representative ensemble algorithms in the literature. 

For feature-based algorithms, their aim is to capture sufficient discriminate features from the given data. For instance, Line \textit{et al.} \cite{b9} introduced a shapelet transformation method to find representative shapelets that reflected the trend of raw data. A bag-of-features representation framework was applied to extracting the information at different locations of sequences \cite{b8}. Dempster \textit{et al.} \cite{b10} applied minimally random convolutional kernel transform to exploring the transformed features from data. In addition, the learned pattern similarity \cite{b11}, bag of symbolic Fourier approximation symbols \cite{b12}, hidden-unit logistic model \cite{b13}, time series forest \cite{b14}, and multi-feature dictionary representation and ensemble learning \cite{b15} are also representative algorithms. 

\subsection{Deep Learning Algorithms}
By unfolding the internal representation hierarchy of data, deep learning algorithms focus on extracting the intrinsic connections among representations. Most of the existing deep learning models are either single-network- or dual-network-based \cite{F12}. A single-network-based model captures the significant correlations within the representation hierarchy of data by one (usually hybridized) network, e.g., FCN \cite{F9}, ResNet \cite{F9}, shapelet-neural network \cite{F13}, InceptionTime \cite{b16}, dynamic temporal pooling \cite{F14}, multi-process collaborative architecture \cite{F31}, and multi-scale attention convolutional neural network \cite{b17}. In contrast, a dual-network-based model usually consists of two parallel networks, i.e., local-feature extraction network (LFN) and global-relation extraction network (GRN), such as FCN-LSTM \cite{F11}, RTFN \cite{F12}, ResNet-Transformer \cite{b18}, RNTS \cite{b19}, and TapNet \cite{F10}.

Almost all algorithms above emphasized single-task TSC, e.g., traffic or gesture classification. However, TSC usually involves multiple tasks in real-world scenarios, like various applications with different TSC tasks run on different mobile devices in a mobile computing environment. Enabling efficient knowledge sharing of similar expertise among different tasks helps increase the average accuracy of these tasks. Nevertheless, sharing knowledge among different TSC tasks securely and efficiently is still a challenge. That is what FL aims for.

\section{EFDLS}
This section first overviews the architecture of EFDLS. Then, it introduces the feature-based student-teacher framework, distance-based weights matching scheme, and communication overhead.

\subsection{System Overview}
EFDLS is a secure distributed system for multi-task TSC. There is a central server and multiple mobile users. Let $N_{tot}$ and $N_{conn}$ denote the numbers of total and connected users in the system, respectively, where $N_{conn} \leq N_{tot}$. Each user runs one TSC task at a time and different users might run different TSC tasks. For two arbitrary users, they run two different tasks, such as gesture and ECG classification, or the same task with different data sources. 

The overview of EFDLS is shown in Fig. 1. In the system, users train their models locally based on knowledge distillation and share their model weights with users with similar expertise via the server. We propose FBST, a feature-based student-teacher framework that is deployed on each user as its learning model. Within each user, its teacher's hidden layers' knowledge is transferred to its student's hidden layers. For each connected user, its student model's hidden layers' weights are uploaded to the EFDLS server periodically. We propose DBWM, a distance-based weights matching scheme deployed on the server, with the LSD adopted to measure the similarity between the weights of two given models. After the weights of all connected users are uploaded completely, for each connected user, the DBWM scheme is launched to find the one with the most similar weights among all connected users. In this way, every user has a partner to match with. For each connected user, its uploaded weights are sent to its partner that then loads these weights to its teacher model's hidden layers. The server's role looks like a telecom-network switch. The EFDLS system allows users to benefit from knowledge sharing without sacrificing security and privacy. 

\subsection{Feature-based Student-Teacher Framework}
In the FBST framework, the student and teacher models have identical network structure. Within each user, feature-based KD promotes knowledge transfer from the teacher’s hidden layers to its student's hidden layers, helping the student capture rich and valuable representations from the input data.

\subsubsection{Feature Extractor}
The feature extractor contains multiple hidden layers and a classifier, as shown in Fig. 1. The hidden layers are responsible for local-feature extraction, including three Convolutional Blocks (i.e., ConvBlock1, ConvBlock2, and ConvBlock3), an average pooling layer, and a dense (i.e., fully-connected) layer. Each ConvBlock consists of a 1-dimensional CNN (Conv) module, a batch normalization (BN) module, and a rectified linear unit activation (ReLU) function, defined as: 
 \begin{equation}
\label{eq7}
\begin{aligned}
f_{convblock} (x) = f_{relu}(f_{bn}(W_{conv} \otimes x + b_{conv}))
\end{aligned}
\end{equation}
where, $W_{conv}$ and $b_{conv}$ are the weight and bias matrices of CNN, respectively. $\otimes$ represents the convolutional computation operation. $f_{bn}$ and $f_{relu}$ denote the batch normalization and ReLU functions, respectively.

Let $x_{bn} = \{x_1,x_2,...,x_{N_{bn}}\}$ denote the input of batch normalization (BN), where $x_{i}$ and $N_{bn}$ stand for the $i$-th instance and batch size, respectively. $f_{bn}(x_{bn})$ is defined in Eq. (\ref{eq8})
 \begin{equation}
\label{eq8}
\begin{aligned}
f_{bn} (x_{bn}) &= f_{bn}(x_1,x_2,...,x_{N_{bn}}) \\
&= (\alpha \frac{x_1-\mu}{\delta +\zeta} + \beta, \alpha \frac{x_2-\mu}{\delta +\zeta} + \beta,...,\alpha \frac{x_{N_{bn}}-\mu}{\delta +\zeta}+ \beta)
\\
 \mu &= \frac{1}{N_{bn}}\sum_{i=1}^{N_{bn}} {x_i}\\
\delta &=  \sqrt{\sum_{i=1}^{N_{bn}}{(x_i-\mu)^{2}}}
\end{aligned}
\end{equation}
where, $\alpha \in \mathbb{R}^{+}$ and $ \beta \in \mathbb{R}$ are the parameters to be learned during training. $\zeta$ $\textgreater$ $0$ is an arbitrarily small number. 

The classifier is composed of a dense layer and a Softmax function, mapping high-level features extracted from the hidden layers to the corresponding label.

\subsubsection{Knowledge Distillation}
Feature-based KD regularizes a student model by transferring knowledge from the corresponding teacher's hidden layers to the student's hidden layers \cite{F29}. For an arbitrary user, its student model captures sufficient discriminate representations from the data under its teacher model's supervision.  

Let $O_{i}^{T,1}$, $O_{i}^{T,2}$, $O_{i}^{T,3}$, and $O_{i}^{T,4}$ be the outputs of ConvBlock 1, ConvBlock 2, ConvBlock 3, and the dense layer of the teacher's hidden layers. Let $O_{i}^{S,1}$, $O_{i}^{S,2}$, $O_{i}^{S,3}$, and $O_{i}^{S,4}$ be the outputs of ConvBlock 1, ConvBlock 2, ConvBlock 3, and the dense layer of the student's hidden layers. Following the previous work \cite{F26}, we define the KD loss, $\mathcal{L}_{i}^{KD}$, of $U_{i}$ as: 
\begin{equation}
\label{eq4}
\begin{aligned}
\mathcal{L}_{i}^{KD} = \sum_{m=1}^{4} ||O_{i}^{T,m}-O_{i}^{S,m}||^{2}
\end{aligned}
\end{equation}

For $U_{i}$, its total loss, $\mathcal{L}_{i}$, consists of KD loss, $\mathcal{L}_{i}^{KD}$, and supervised loss, $\mathcal{L}_{i}^{Sup}$. As the previous studies in \cite{F10,F11,F12} suggested, $\mathcal{L}_{i}^{Sup}$ uses the cross-entropy function to measure the average difference between the ground truth labels and their prediction vectors, as shown in Eq. (\ref{eq5}).
\begin{equation}
\label{eq5}
\begin{aligned}
\mathcal{L}_{i}^{Sup} = -\frac{1}{N_{seg}} \sum_{j=1}^{N_{seg}}{y_{j}log(\hat{y}_{j})}
\end{aligned}
\end{equation}
where, $N_{seg}$ is the number of input vectors, and $y_{i}$ and $\hat{y}_j$ are the ground label and prediction vector of the $j$-th input vector, respectively.

The total loss, $\mathcal{L}_{i}$, is defined as:
\begin{equation}
\label{eq6}
\begin{aligned}
\mathcal{L}_{i} = \epsilon \times \mathcal{L}_{i}^{Sup} + (1-\epsilon) \times \mathcal{L}_{i}^{KD}
\end{aligned}
\end{equation}
where, $\epsilon \in (0,1)$ is a coefficient to balance $\mathcal{L}_{i}^{Sup}$ and $\mathcal{L}_{i}^{KD}$. In this paper, we set $\epsilon$ = 0.9 (More details can be found in Section 4.3). 

\subsection{Distance-based Weights Matching}
The least square distance (LSD) is used to calculate the similarity between the weights of two given models. When the weights uploaded by all the connected users are received, the DBWM scheme immediately launches the weights matching process to find a partner for each connected user. 

\subsubsection{Least Square Distance} 
Let $FLEs$ denote the maximum number of federated learning epochs. Let $W_{i}^{S,k}$ and $W_{i}^{T,k}$ be the weights of the student and teacher models of $U_{i}$ at the $k$-th federated learning epoch, $k = 1,2,...,FLEs$. Denote the hidden layers' weights of the student and teacher models of $U_{i}$ by $W_{i}^{S_{hidden},k} \subset  W_{i}^{S,k}$ and $W_{i}^{T_{hidden},k} \subset  W_{i}^{T,k}$, respectively. To be specific, $W_{i}^{S_{hidden},k}$ consists of the weights of ConvBlock 1, ConvBlock 2, ConvBlock 3, and the dense layer, namely, $W_{i}^{S_{1},k}$, $W_{i}^{S_2,k}$, $W_{i}^{S_3,k}$, and $W_{i}^{S_4,k}$. So, we have $W_{i}^{S_{hidden},k} = \{W_{i}^{S_1,k},W_{i}^{S_2,k},$ $ W_{i}^{S_3,k}, W_{i}^{S_4,k}\}$. 
 
At the $k$-th federated learning epoch, user $U_{i}, i = 1,2,...,N_{conn}$, uploads its student model's hidden layers' weights, $W_{i}^{S_{hidden},k}$, to the server. The server stores the uploaded weights in the weight set $\textbf{W}$ defined in Eq. (\ref{deq1}).
 \begin{equation}
\label{deq1}
\begin{aligned}
\textbf{W} = [W_{1}^{S_{hidden},k},W_{2}^{S_{hidden},k},...,W_{N_{conn}}^{S_{hidden},k}]
\end{aligned}
\end{equation}

The server then calculates the weights' square distance set, $d$, based on $\textbf{W}$. $d$ is defined as:
\begin{equation}       
\label{deq2}
d=\left[                 %左括号
  \begin{array}{c}   %该矩阵一共3列，每一列都居中放置
    d_{1} \\  %第一行元素
    d_{2} \\  %第二行元素
    ...\\
    d_{N_{conn}}
  \end{array}
\right]=        \left[                 %左括号
  \begin{array}{ccc}   
    d_{1,2}&...&d_{1,N_{conn}} \\  %第一行元素
     d_{2,1}&...&d_{2,N_{conn}} \\  
    ...&...&...\\
    d_{N_{conn},1}&...&d_{N_{conn},N_{conn} - 1}
  \end{array}
\right]         
\end{equation}
where, $d_{i,j}$ ($i,j \in {1,...,N_{conn}}, i \neq j$) is the square distance between $W_{i}^{S_{hidden},k}$ and $W_{j}^{S_{hidden},k}$, as defined in Eq. (\ref{eq1}).
 \begin{equation}
\label{eq1}
\begin{aligned}
d_{i,j} &= ||W_{i}^{S_{hidden},k}-W_{j}^{S_{hidden},k}||^{2} \\
&= \sum_{m=1}^{4}{||W_{i}^{S_m,k}-W_{j}^{S_m,k}||^2}  
\end{aligned}
\end{equation}
We adopt the $argmin$ function to return the index of the smallest distance for each row in $d$ and obtain the index set, $\textbf{ID}$. $\textbf{ID}$ is defined in Eq. (\ref{deq3}).
\begin{equation}
\label{deq3}
\begin{aligned}
\textbf{ID} = argmin(d) = [ID_1,ID_2,...,ID_{N_{conn}}]
\end{aligned}
\end{equation}
where, $ID_i$ is the index of the smallest distance for $U_i$.

Based on $\textbf{ID}$, we easily obtain the LSD weight set, $\textbf{W}^{LSD}$, from $\textbf{W}$. $\textbf{W}^{LSD}$ is defined in Eq. (\ref{deq4}).
 \begin{equation}
\label{deq4}
\begin{aligned}
\textbf{W}^{LSD} &= [W_{1}^{LSD,k},W_{2}^{LSD,k},...,W_{N_{conn}}^{LSD,k}]\\
&= [ \textbf{W}(ID_{1}), \textbf{W}(ID_{2}),..., \textbf{W}(ID_{N_{conn}})]
  \end{aligned}
\end{equation}
where, $W_{i}^{LSD,k}$ are the weights matched with those of $U_i$ at the $k$-th federated learning epoch.

Once $U_{i}$ receives $W_{i}^{LSD,k}$ from the server, $U_{i}$ loads these weights to its teacher's hidden layers at the beginning of the next federated learning epoch, as defined in Eq. (\ref{deq5}).
 \begin{equation}
\label{deq5}
\begin{aligned}
W_{i}^{T_{hidden},k+1} \leftarrow W_{i}^{LSD,k} 
\end{aligned}
\end{equation}

Alg. 1 and Alg. 2 show the user and server implementation procedures, respectively.

 \begin{algorithm}[t]
\caption{EFDLS User Implementation Procedure}
\begin{algorithmic}[1]
\label{ag1}
\Procedure{UserProcedure}{$U_i,FLEs$}       %\Comment{N, H and C: the number of users, existing users and epochs, respectively}
    \State Initialize all global variables;
    \For {$k=1$ to $FLEs$}  %\Comment{Users first upload their weights }
    \If {$k == 1$} 
    \State // The student model is trained alone
    \State Obtain $W_{i}^{S,k}$ after the initial local training;
    \State // Upload its hidden layers' weights to server
    \State Upload $W_{i}^{S_{hidden},k} \subset W_{i}^{S,k}$; 
    \Else
    \If{receiveServer(Active)==1}
    \State // Connect to the EFDLS server
    \State Receive $W_{i}^{LSD,k}$;
    \State Load $W_{i}^{LSD,k}$ to the teacher model;
    \State Compute $\mathcal{L}_{i}$ by Eq. (\ref{eq6});
    \State Update $W_{i}^{S,k+1}$ using the gradient decent; 
    \State Upload $W_{i}^{S_{hidden},k+1} \subset W_{i}^{S,k+1}$; 
    \Else
     \State Disconnect from the EFDLS server.
    \EndIf 
   \EndIf
    \EndFor
    %\State \Return $w_{H}^{C}$ \Comment{weights of model of the last user}
\EndProcedure

\end{algorithmic}
\end{algorithm}
 
 \begin{algorithm}[t]
\caption{EFDLS Server Implementation Procedure}
\begin{algorithmic}[1]
\label{ag2}
\Procedure{ServerProcedure}{$N_{tot},N_{conn},FLEs$}       
   \State Initialize all global variables;
    \State Set $ \textbf{W} =\varnothing$;
    %\State Set $d = \varnothing$;
    \For {$k=1$ to $FLEs$}  %\Comment{Users first upload their weights }
    \State // Run on the server;
    \State Clear and initialize $\textbf{W}$;
        \For {$i=1$ to $N_{conn}$}
        \State // Receive model weights from users;
       	\State Receive $W_{i}^{S_{hidden},k}$; 
	\State Include $W_{i}^{S_{hidden},k}$ in $\textbf{W}$.
        \EndFor
     \For {$i=1$ to $N_{conn}$}
     \State Obtain $W_{i}^{LSD,k}$ based on $\textbf{W}$ by Eqs. (\ref{deq1})-(\ref{deq4});
     \State  Send $W_{i}^{LSD,k}$ to $U_{i}$.
    \EndFor
    \EndFor
    %\State \Return $w_{H}^{C}$ \Comment{weights of model of the last user}
\EndProcedure

\end{algorithmic}
\end{algorithm}

\begin{table*}[t]
  \caption{Details of 44 selected datasets from the UCR 2018.}
  \label{table1}
  \centering
  \resizebox{130mm}{80mm}{
\label {dd}
\begin{tabular}{ccccccc}
\hline
Scale                    & Dataset              & Train & Test & Class & SeriesLength & Type       \\ \hline
\multirow{11}{*}{Short}  & Chinatown            & 20    & 345  & 2     & 24           & Traffic    \\
                         & MelbournePedestrian  & 1194  & 2439 & 10    & 24           & Traffic    \\
                         & SonyAIBORobotSur.2   & 27    & 953  & 2     & 65           & Sensor     \\
                         & SonyAIBORobotSur.1   & 20    & 601  & 2     & 70           & Sensor     \\
                         & DistalPhalanxO.A.G   & 400   & 139  & 3     & 80           & Image      \\
                         & DistalPhalanxO.C.    & 600   & 276  & 2     & 80           & Image      \\
                         & DistalPhalanxTW      & 400   & 139  & 6     & 80           & Image      \\
                         & TwoLeadECG           & 23    & 1139 & 2     & 82           & ECG        \\
                         & MoteStrain           & 20    & 1252 & 2     & 84           & Sensor     \\
                         & ECG200               & 100   & 100  & 2     & 96           & ECG        \\
                         & CBF                  & 30    & 900  & 3     & 128          & Simulated  \\ \hline
\multirow{11}{*}{Medium} & DodgerLoopDay        & 78    & 80   & 7     & 288          & Sensor     \\
                         & DodgerLoopGame       & 20    & 138  & 2     & 288          & Sensor     \\
                         & DodgerLoopWeekend    & 20    & 138  & 2     & 288          & Sensor     \\
                         & CricketX             & 390   & 390  & 12    & 300          & Motion     \\
                         & CricketY             & 390   & 390  & 12    & 300          & Motion     \\
                         & CricketZ             & 390   & 390  & 12    & 300          & Motion     \\
                         & FaceFour             & 24    & 88   & 4     & 350          & Image      \\
                         & Ham                  & 109   & 105  & 2     & 431          & Spectro    \\
                         & Meat                 & 60    & 60   & 3     & 448          & Spectro    \\
                         & Fish                 & 175   & 175  & 7     & 463          & Image      \\
                         & Beef                 & 30    & 30   & 5     & 470          & Spectro    \\ \hline
\multirow{11}{*}{Long}   & OliveOil             & 30    & 30   & 4     & 570          & Spectro    \\
                         & Car                  & 60    & 60   & 4     & 577          & Sensor     \\
                         & Lightning2           & 60    & 61   & 2     & 637          & Sensor     \\
                         & Computers            & 250   & 250  & 2     & 720          & Device     \\
                         & Mallat               & 55    & 2345 & 8     & 1024         & Simulated  \\
                         & Phoneme              & 214   & 1896 & 39    & 1024         & Sensor     \\
                         & StarLightCurves      & 1000  & 8236 & 3     & 1024         & Sensor     \\
                         & MixedShapesRegularT. & 500   & 2425 & 5     & 1024         & Image      \\
                         & MixedShapesSmallT.   & 100   & 2425 & 5     & 1024         & Image      \\
                         & ACSF1                & 100   & 100  & 10    & 1460         & Device     \\
                         & SemgHandG.Ch2        & 300   & 600  & 2     & 1500         & Spectrum   \\ \hline
\multirow{11}{*}{Vary}   & AllGestureWiimoteX   & 300   & 700  & 10    & Vary         & Sensor     \\
                         & AllGestureWiimoteY   & 300   & 700  & 10    & Vary         & Sensor     \\
                         & AllGestureWiimoteZ   & 300   & 700  & 10    & Vary         & Sensor     \\
                         & GestureMidAirD1      & 208   & 130  & 26    & Vary         & Trajectory \\
                         & GestureMidAirD2      & 208   & 130  & 26    & Vary         & Trajectory \\
                         & GestureMidAirD3      & 208   & 130  & 26    & Vary         & Trajectory \\
                         & GesturePebbleZ1      & 132   & 172  & 6     & Vary         & Sensor     \\
                         & GesturePebbleZ2      & 146   & 158  & 6     & Vary         & Sensor     \\
                         & PickupGestureW.Z     & 50    & 50   & 10    & Vary         & Sensor     \\
                         & PLAID                & 537   & 537  & 11    & Vary         & Device     \\
                         & ShakeGestureW.Z      & 50    & 50   & 10    & Vary         & Sensor     \\ \hline
\end{tabular}
}
\end{table*}
\subsection{Communication Overhead}
EFDLS does not launch the DBWM scheme unless the weights from all the $N_{conn}$ connected users are received. It helps reduce the interaction between the server and users, promoting the system's service efficiency. For user $U_{i}, i = 1, 2, ..., N_{conn}$, we analyze the communication overhead of uploading and downloading its weights. Denote the bandwidth requirement for uploading the student model's hidden layers' weights of $U_{i}$ once by $BW$. Clearly, the bandwidth requirement for downloading the student model's hidden layers' weights from the server once is also $BW$. That is because, for an arbitrary connected user, the weights uploaded to and those downloaded from the server are of the same size, given that each user has exactly the same model structure. At each federated learning epoch, the bandwidth requirement for user $U_{i}, i = 1, 2, ..., N_{conn}$ is estimated as $BW + BW = 2BW$. For $U_{i}$, its total communication overhead is in proportion to $2BW \cdot FLEs$. Hence, the total communication overhead is proportional to $2BW \cdot FLEs \cdot N_{conn}$.

\section{Performance Evaluation}
This section first introduces the experimental setup and performance metrics and then focuses on the ablation study. Finally, the performance of EFDLS is evaluated.
\subsection{Experimental Setup}
\subsubsection{Data Description}
The UCR 2018 archive is one of the most popular time series repositories with 128 datasets of different lengths in various application domains \cite{F30}. Following the previous work \cite{F31}, we divide the UCR 2018 archive into 4 categories with respect to dataset length, namely, `short’, `medium’, `long’, and `vary’. The length of a `short’ dataset is no more than 200. That of a `medium’ one varies from 200 to 500. A `long’ one has a length of over 500 while a `vary’ one has an indefinite length. The 128 datasets are composed of 41 `short’ , 32 `medium’, 44 `long', and 11 `vary’ datasets. Unfortunately, our limited computing resources do not allow us to consider the whole 128 datasets (detailed hardware specification can be found in Subsection \textit{Implementation Details}). There are seven algorithms for performance comparison and the average training time on the 128 datasets costed more than 32 hours for a single federated learning epoch. So, we select 11 datasets from each category, resulting in 44 datasets selected. More details are found in Table \ref{table1}. 

\begin{table*}[t]
  \caption{Experimental results of different algorithms on 44 datasets when $N_{conn}$ = 44 and $N_{tot}$ = 44.}
  \label{table2}
  \centering
  \resizebox{150mm}{93.8mm}{
\begin{tabular}{ccccccccc}
\hline
Dataset              & Baseline        & FedAvg & FedAvgM & FedGrad & FTL             & FTLS            & FKD            & EFDLS            \\\hline
Chinatown            & 0.9623          & 0.2754 & 0.2754  & 0.9623      & \textbf{0.9665} & 0.9537          & 0.9275          & 0.9478          \\
MelbournePedestrian  & 0.9139          & 0.1    & 0.1     & 0.7784      & 0.8486          & 0.8922          & 0.9379          & \textbf{0.9453} \\
SonyAIBORobotSur.2   & 0.8961          & 0.383  & 0.383   & 0.8363      & 0.8688          & 0.9035          & \textbf{0.915}  & 0.8961          \\
SonyAIBORobotSur.1   & 0.8652          & 0.5707 & 0.6619  & 0.7887      & 0.8236          & 0.8702          & 0.8369          & \textbf{0.8819} \\
DistalPhalanxO.A.G   & \textbf{0.6763} & 0.1079 & 0.1079  & 0.6187      & 0.6259          & 0.6475          & 0.6691          & 0.6475          \\
DistalPhalanxO.C.    & 0.75            & 0.417  & 0.6619  & 0.6776      & 0.7464          & 0.7465          & \textbf{0.7536} & 0.7428          \\
DistalPhalanxTW      & 0.6547          & 0.1295 & 0.1295  & 0.554       & 0.6259          & 0.6547          & \textbf{0.6835} & 0.6403          \\
TwoLeadECG           & 0.7463          & 0.4996 & 0.4996  & 0.7305      & 0.7287          & 0.7278          & \textbf{0.8112} & 0.7665          \\
MoteStrain           & 0.7788          & 0.5391 & 0.5391  & 0.6933      & 0.7923          & \textbf{0.8283} & 0.8163          & 0.8203          \\
ECG200               & 0.86            & 0.36   & 0.36    & 0.8         & 0.84            & 0.85            & \textbf{0.87}   & 0.85            \\
CBF                  & 0.987           & 0.3333 & 0.5911  & 0.5911      & 0.973           & 0.9922          & 0.9922          & \textbf{0.9956} \\
DodgerLoopDay        & \textbf{0.575}  & 0.15   & 0.15    & 0.3875      & 0.55            & 0.525           & 0.5125          & 0.5375          \\
DodgerLoopGame       & 0.6884          & 0.5217 & 0.5217  & 0.6232      & \textbf{0.7826} & 0.7609          & 0.7609          & 0.7464          \\
DodgerLoopWeekend    & 0.8261          & 0.7391 & 0.7391  & 0.7319      & 0.8841          & 0.8913          & 0.913           & \textbf{0.9203} \\
CricketX             & 0.5897          & 0.0692 & 0.1371  & 0.2256      & 0.5667          & 0.6128          & 0.659           & \textbf{0.6718} \\
CricketY             & 0.5051          & 0.0949 & 0.1357  & 0.1949      & 0.5             & 0.4949          & 0.5538          & \textbf{0.5974} \\
CricketZ             & 0.6205          & 0.0846 & 0.0846  & 0.2256      & 0.5692          & 0.6             & 0.6692          & \textbf{0.7256} \\
FaceFour             & 0.6477          & 0.1591 & 0.1591  & 0.4659      & 0.6591          & \textbf{0.6932} & \textbf{0.6932} & 0.6818          \\
Ham                  & \textbf{0.7143} & 0.4857 & 0.4857  & 0.6762      & 0.7048          & \textbf{0.7143} & 0.7048          & 0.6952          \\
Meat                 & 0.8667          & 0.3333 & 0.3333  & 0.7333      & 0.8333          & 0.8333          & 0.9             & \textbf{0.917}  \\
Fish                 & 0.5657          & 0.1371 & 0.1371  & 0.2857      & 0.5771          & 0.6             & 0.6             & \textbf{0.6229} \\
Beef                 & \textbf{0.7667} & 0.2    & 0.2     & 0.5667      & 0.7             & 0.7             & 0.7             & \textbf{0.7667} \\
OliveOil             & 0.8333          & 0.167  & 0.167   & 0.7         & \textbf{0.8667} & \textbf{0.8667} & 0.8333          & 0.8333          \\
Car                  & 0.5833          & 0.233  & 0.233   & 0.5         & 0.5667          & 0.5833          & 0.5667          & \textbf{0.6333} \\
Lightning2           & 0.7869          & 0.459  & 0.459   & 0.7705      & 0.7869          & \textbf{0.8033} & 0.7541          & 0.7869          \\
Computers            & 0.78            & 0.5    & 0.5     & 0.584       & 0.688           & 0.748           & 0.788           & \textbf{0.804}  \\
Mallat               & 0.7446          & 0.1254 & 0.1254  & 0.4141      & 0.7638          & 0.7539          & 0.7906          & \textbf{0.8299} \\
Phoneme              & 0.2231          & 0.02   & 0.02    & 0.1108      & 0.2147          & 0.2247          & 0.2859          & \textbf{0.2954} \\
StarLightCurves      & 0.9534          & 0.1429 & 0.1429  & 0.5062      & 0.9519          & \textbf{0.9584} & 0.9571          & 0.9582          \\
MixedShapesRegularT. & 0.8586          & 0.1889 & 0.1889  & 0.2223      & 0.8384          & 0.8598          & 0.8643          & \textbf{0.8907} \\
MixedShapesSmallT.   & 0.8029          & 0.1889 & 0.1889  & 0.2421      & 0.7942          & 0.8062          & 0.8318          & \textbf{0.8388} \\
ACSF1                & 0.77            & 0.1    & 0.19    & 0.19        & 0.82            & \textbf{0.89}   & 0.87            & 0.88            \\
SemgHandG.Ch2        & 0.7067          & 0.65   & 0.65    & 0.555       & 0.72            & \textbf{0.7383} & 0.6867          & 0.72            \\
AllGestureWiimoteX   & 0.2643          & 0.1    & 0.1     & 0.1371      & 0.2729          & \textbf{0.3043} & 0.2929          & 0.2914          \\
AllGestureWiimoteY   & 0.2585          & 0.1    & 0.1     & 0.1357      & \textbf{0.3186} & 0.3029          & 0.2529          & 0.2829          \\
AllGestureWiimoteZ   & 0.2886          & 0.1    & 0.1     & 0.1343      & 0.2671          & 0.29            & \textbf{0.4014} & 0.3786          \\
GestureMidAirD1      & 0.5538          & 0.0384 & 0.0384  & 0.0923      & 0.5462          & 0.5538          & 0.4615          & \textbf{0.5769} \\
GestureMidAirD2      & 0.4231          & 0.0384 & 0.0384  & 0.0923      & 0.4154          & 0.4462          & 0.4692          & \textbf{0.5308} \\
GestureMidAirD3      & \textbf{0.3}    & 0.0384 & 0.0384  & 0.0923      & 0.2693          & 0.2615          & 0.2231          & 0.2769          \\
GesturePebbleZ1      & 0.4419          & 0.1628 & 0.1628  & 0.2558      & 0.4767          & 0.4826          & \textbf{0.5}    & 0.4883          \\
GesturePebbleZ2      & 0.4241          & 0.1519 & 0.1519  & 0.2722      & 0.5126          & 0.557           & \textbf{0.6013} & 0.5886          \\
PickupGestureW.Z     & 0.56            & 0.1    & 0.1     & 0.24        & 0.62            & 0.6             & 0.7             & \textbf{0.74}   \\
PLAID                & 0.203           & 0.0615 & 0.0615  & 0.0615      & 0.2198          & 0.2253          & \textbf{0.2924} & 0.2589          \\
ShakeGestureW.Z      & 0.92            & 0.1    & 0.1     & 0.1         & \textbf{0.96}   & 0.92            & \textbf{0.96}   & \textbf{0.96}   \\\hline
Win                  & 4               & 0      & 0       & 0           & 3               & 7               & 10              & \textbf{18}     \\
Tie                 & 1               & 0      & 0       & 0           & \textbf{2}               & 1               & 1               & \textbf{2}      \\
Lose                  & 39              & 44     & 44      & 44          & 39              & 36              & 33              & \textbf{24}     \\
Best           &5&0&0&0&5&8&11&\textbf{20}\\
MeanACC              & 0.6622          & 0.2377 & 0.2557  & 0.4445      & 0.6604          & 0.6743          & 0.6878          & \textbf{0.7014} \\
AVG\_rank            & 3.5455          & 7.5    & 7.3409  & 6.0113      & 3.9204          & 2.8977          & 2.6364          & \textbf{2.1478}\\\hline

\end{tabular}  
}
\end{table*}

\subsubsection{Implementation Details}
Following previous studies \cite{F8,F9,F10,F11,F31}, we set the decay value of batch normalization to 0.9. We use the $L_2$ regularization to avoid overfitting during the training process. Meanwhile, we adopt the AdamOptimizer with Pytorch\footnote{\url{https://pytorch.org/}}, where the initial learning rate is set to 0.0001. 

All experiments were conducted on a desktop with an Nvidia GTX 1080Ti GPU with 11GB memory, and an AMD R5 1400 CPU with 16G RAM under the Ubuntu 18.04 OS.
 
\subsection{Performance Metrics}
To evaluate FL algorithms' performance, we use three well-known metrics: `win’/`tie’/`lose', mean accuracy (MeanACC), and AVG\_rank, all based on the top-1 accuracy. For an arbitrary algorithm, its `win’, `tie’, and `lose’ values indicate on how many datasets it is better than, equal to, and worse than the others, respectively; its `best' value is the summation of the corresponding `win' and `tie' values. The AVG\_rank score reflects the average difference between the accuracy values of a model and the best accuracy values among all models \cite{F9,F10,F11,F12,b19}.

 \begin{figure}[t]
\label{fi2}
  \centering
\includegraphics[width=9cm]{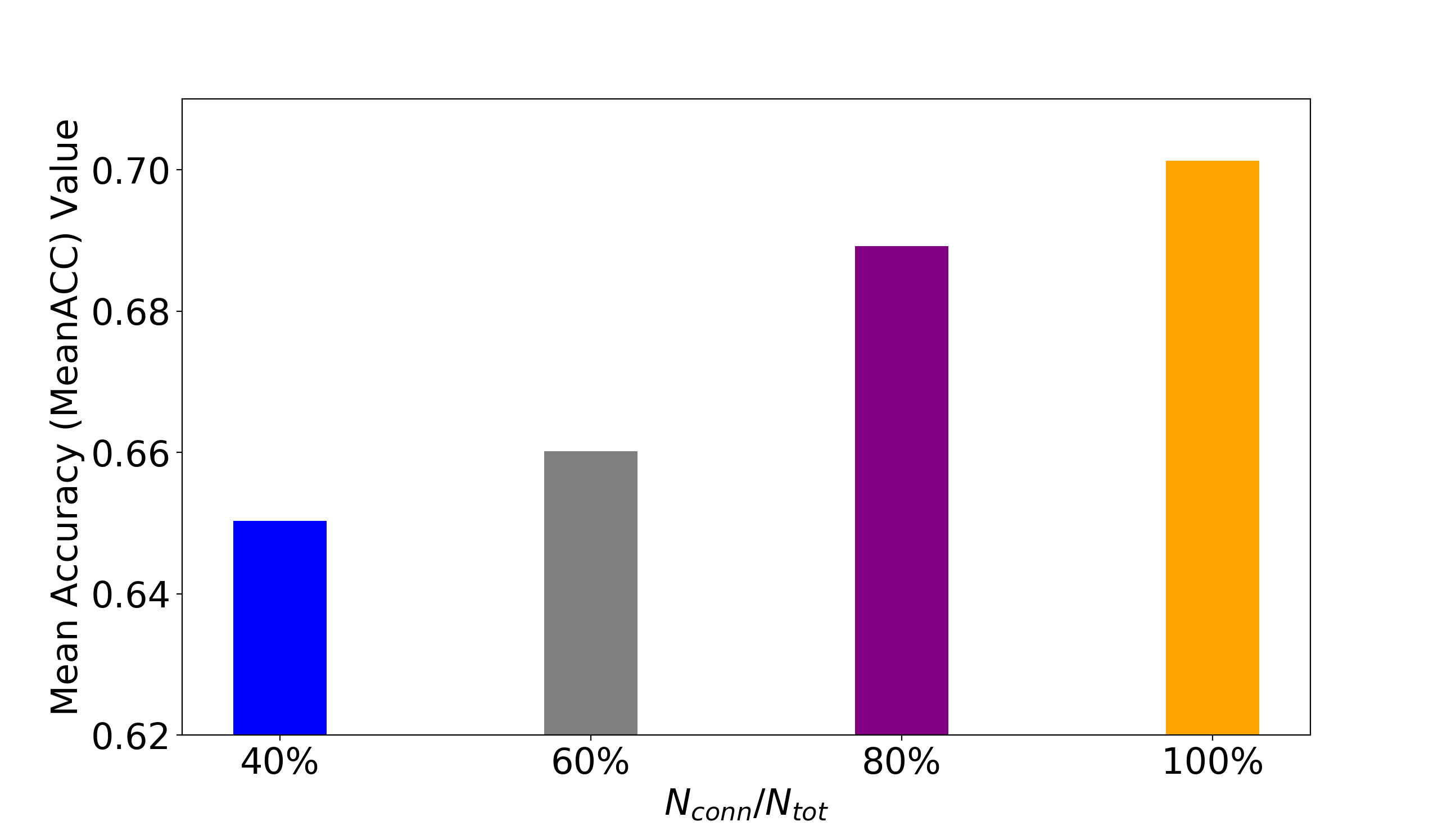}
\caption{MeanACC results obtained by EFDLS with different ratios of $N_{conn}$ to $N_{tot}$ on 44 datasets when $N_{tot}$ = 44.}
\end{figure}
 
\begin{figure}[t]
\label{fi3}
  \centering
\includegraphics[width=9cm]{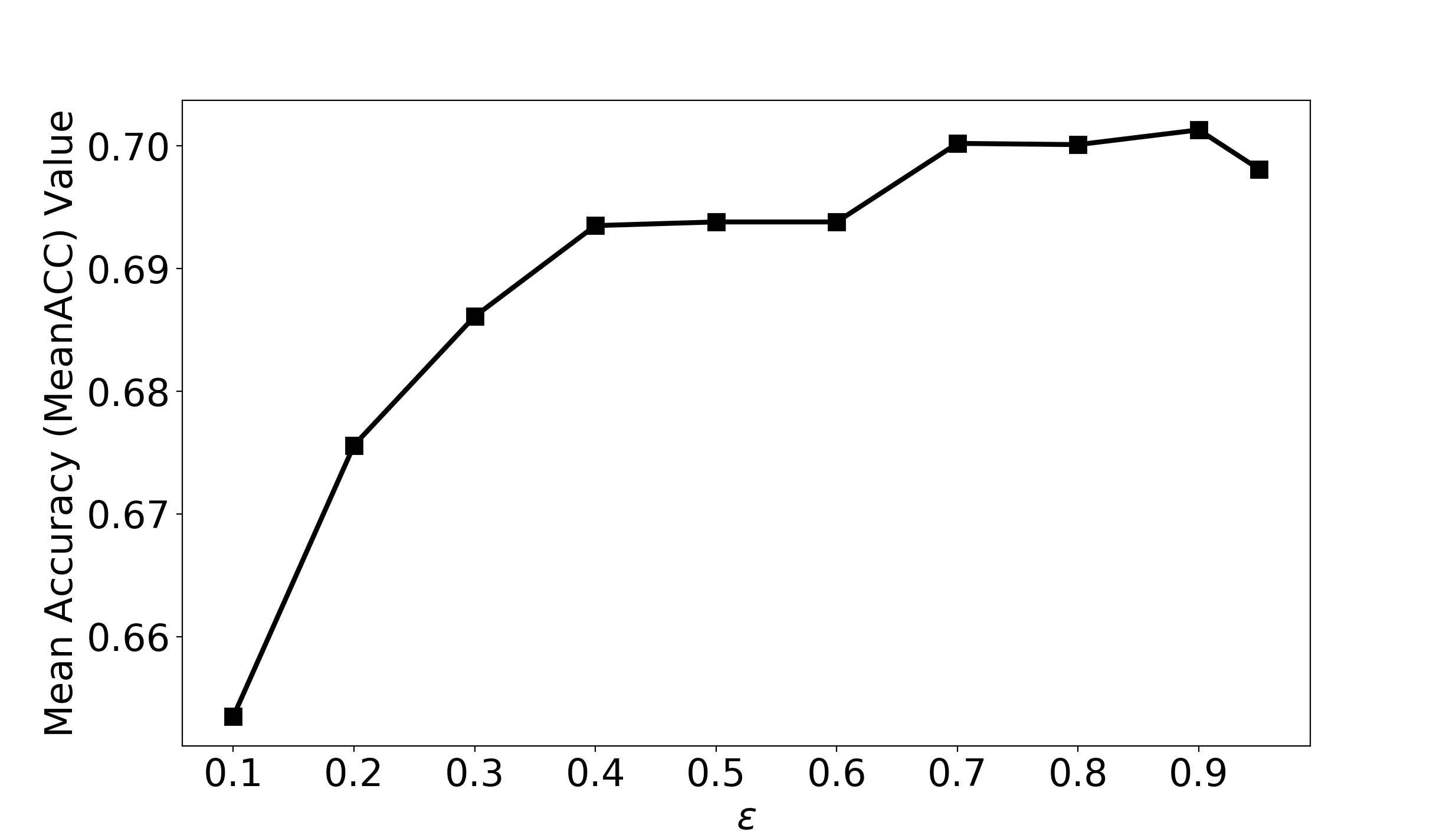}
\caption{MeanACC results with different $\epsilon$ values on 44 datasets when $N_{conn}$ = 44 and $N_{tot}$ = 44.}
\end{figure}

\subsection{Ablation Study}
We use the 44 UCR2018 datasets above to study the impact of parameter settings on the performance of EFDLS. Assume there are 44 users in the system, i.e., $N_{tot} = 44$. Each user runs a TSC task with data coming from a specific dataset. For any two users, if they run identical tasks, e.g., motion recognition, their data sources come from different datasets, e.g., CricketX and CricketY. In the experiments, each user's data comes from one of the 44 datasets.

\subsubsection{Impact of $N_{conn}$}
To investigate the impact of $N_{conn}$ on the EFDLS's performance, we select four ratios of $N_{conn}$ to $N_{tot}$, namely 40\%, 60\%, 80\%, and 100\%. For example, 40\% means there are 18 connected users for weights uploading, given $N_{tot}$ = 44. The MeanACC results obtained by EFDLS with different $N_{conn}$ values on 44 datasets are shown in Fig. 2. One can easily observe that a larger $N_{conn}$ tends to result in a higher MeanACC value. That is because as $N_{conn}$ increases, more amount of time series data is made use of by the system and thus more discriminate representations are captured.

\subsubsection{Impact of $\epsilon$}
$\epsilon$ is a coefficient to balance each connected user's supervised and KD losses in EFDLS. Fig. 3 shows the MeanACC results with different $\epsilon$ values when $N_{conn}$ = 44 and $N_{tot}$ = 44. It is seen that $\epsilon$ = 0.90 results in the highest MeanACC score, i.e., 0.7014. That means $\epsilon$ = 0.90 is appropriate to reduce each user's entropy on its data during training.

\begin{figure*}[t]

  \centering
\includegraphics[width=1.0\textwidth]{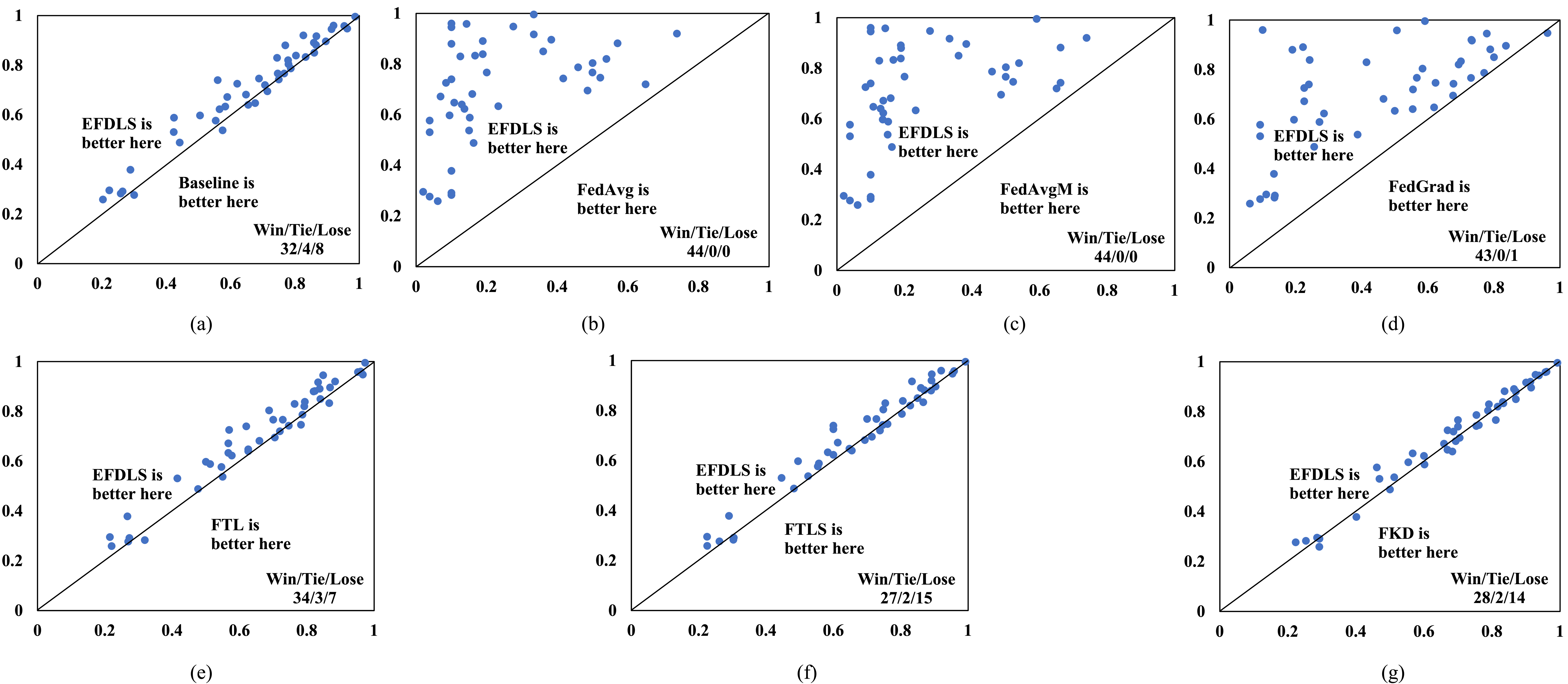}
\caption{ Accuracy plot results showing the performance difference between two given algorithms. (a) Baseline vs. EFDLS; (b) FedAvg vs. EFDLS; (c) FedAvgM vs. EFDLS; (d) FedGrad vs. EFDLS; (e) FTL vs. EFDLS; (f) FTLS vs. EFDLS; (g) FKD vs. EFDLS.
}
\label{fi6}
\end{figure*} 

\subsection{Experimental Analysis}
To evaluate the overall performance of EFDLS, we compare it with seven benchmark algorithms listed below against `Win’/`Lose'/`Tie', MeanACC, and AVG\_rank.
\begin{itemize}
\item Baseline: the single-task TSC algorithm with the feature extractor in Fig. 1 deployed on each user. Note that each user has a unique dataset to run and knowledge sharing among them is disabled.
\item FedAvg: the FederatedAveraging method using the feature extractor in Fig. 1\cite{F17}. 
\item FedAvgM: the modified FedAvg using the feature extractor in Fig. 1 \cite{F25}.
\item FedGrad: the federated gradient method using the feature extractor in Fig. 1 \cite{F16}.
\item FTL: the federated transfer learning method using the feature extractor in Fig. 1 \cite{F22}.
\item FTLS: FTL \cite{F22} based on the DBWM scheme using the feature extractor in Fig. 1.
\item FKD: the federated knowledge distillation using the feature extractor in Fig. 1 \cite{F25,F26}. For fair comparison, FKD uses the same student-teacher network structure as EFDLS.
\end{itemize}

\begin{figure*}[t]
\centering
\includegraphics[width=1.0\textwidth]{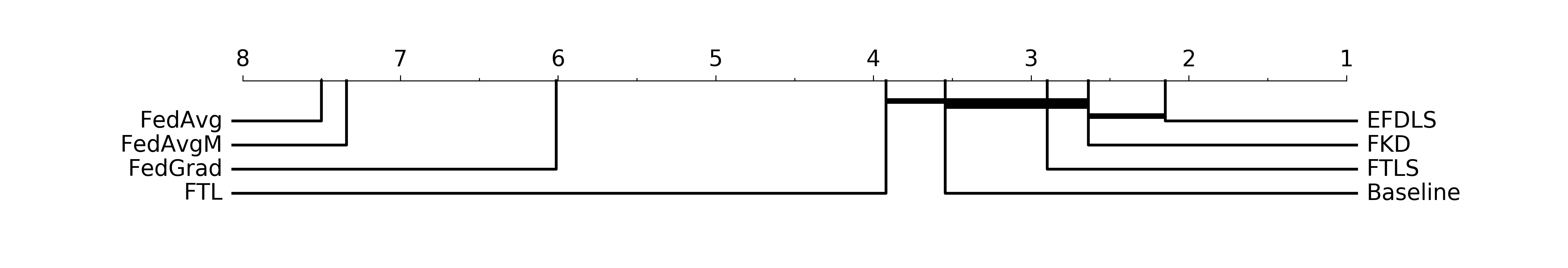} % Reduce the figure size so that it is slightly narrower than the column.
\caption{Critical difference diagram of the average ranks of various FL algorithms on 44 datasets.}
\label{fig5}
\end{figure*}

Table \ref{table2} shows the top-1 accuracy results with various algorithms on 44 UCR2018 datasets when $N_{conn}$ = 44 and $N_{tot}$ = 44. To visualize the differences between EFDLS and the others, Fig. 4 depicts the accuracy plots of EFDLS against each of the remaining algorithms on 44 datasets. In addition, the AVG\_rank results are shown in Fig. \ref{fig5}.

First of all, we study the effectiveness of \textit{knowledge sharing among users} by comparing EFDLS with Baseline. One can observe that EFDLS beats Baseline in every aspect, including `Win’/`Lose'/`Tie', MeanACC, and AVG\_rank. For example, the former wins 18 out of 44 datasets while the latter wins only 4. The accuracy plot of EFDLS vs. Baseline in Fig. 4(a) also supports the finding above. The main difference between EFDLS and Baseline is that the latter only uses standalone feature extractors which do not share the locally collected knowledge with each other. On the other hand, with sufficient knowledge sharing of similar expertise among users enabled, EFDLS improves the system's generalization ability and thus achieves promising multi-task TSC performance. 

Secondly, we study the effectiveness of \textit{the FBST framework} by comparing EFDLS with FTLS. It is easily seen that EFDLS outperforms FTLS regarding the `best', MeanACC, and AVG\_rank values. The accuracy plot of EFDLS vs. FTLS in Fig. 4(f) also supports this. The FBST framework allows efficient knowledge transfer from teacher to student, helping the student capture sufficient discriminate representations from the input data. On the contrary, the FTLS's learning model lacks of self-generalization, leading to deteriorated performance during knowledge sharing. 

Thirdly, we study the effectiveness of \textit{the DBWM scheme} by comparing EFDLS with FKD. Apparently, EFDLS overweighs FKD with respect to `best', MeanACC, and AVG\_rank. It is backed by the accuracy plot of EFDLS vs. FTLS in Fig. 4(g). As mentioned before, at each federated learning epoch, the DBWM scheme finds a partner for each user and then EFDLS offers weights exchange between each pair of connected users, which realizes knowledge sharing of similar expertise among different users. In contrast, FKD adopts the average weights to supervise the feature extraction process in each user. It is likely to lead to catastrophic forgetting in a user whose weights significantly differ from the average weights.

Last but not least, we compare EFDLS with all the seven algorithms. One can easily observe that our EFDLS is no doubt the best among all algorithms for comparison since ours obtains the highest MeanACC and `best’ values, namely 0.7014 and 20, and the smallest AVG\_rank value, namely 2.1478. The FKD takes the second position when considering its `best', MeanACC, and AVG\_rank values, namely, 11, 0.6878, and 2.6364. On the other hand, FedAvg and its variant, FedAvgM, are the two worst algorithms. The following explains the reasons behind the findings above. When faced with the multi-task TSC problem, each user runs one TSC task, and different users may run different TSC tasks. The FBST framework and the DBWM scheme help EFDLS to realize fine-grained knowledge sharing between any pair of users with the most similar expertise. FKD uses the average of all users' weights to guide each user to capture valuable features from the data, promoting coarse-grained knowledge sharing among users. On the other hand, FedAvg and FedAvgM simply take the average weights of all users as each user's weights, which may cause catastrophic forgetting and hence poor performance on multi-task TSC. 

\section{Conclusion}

The FBST framework promotes knowledge transfer from a teacher's to its student's hidden layers, helping the student capture instance-level representations from the input. The DBWM scheme finds a partner for each user in terms of similarity between their uploaded weights, enabling knowledge sharing of similar expertise among different users. With FBST and DBWM, the proposed EFDLS securely shares knowledge of similar expertise among different tasks for multi-task time series classification. Experimental results show that compared with six benchmark FL algorithms, EFDLS is a winner on 44 datasets with respect to the MeanACC and AVG\_rank metrics and on 20 datasets in terms of the `best' measure. In particular, compared with single-task Baseline, EFDLS obtains 32/4/8 regarding the `win'/`tie'/`lose' metric. That reflects the potential of EFDLS to be applied to multi-task TSC problems in various real-world domains.

%
%\appendices


%\ifCLASSOPTIONcompsoc
  %\section*{Acknowledgments}
%\else
%\fi


\ifCLASSOPTIONcaptionsoff
  \newpage
\fi



%\bibliographystyle{IEEEtran}

\normalem
\bibliographystyle{IEEEtran} 
\bibliography{IEEEexample}

%\end{IEEEbiography}


%\newpage


\end{document}